\pdfoutput=1
\documentclass[11pt]{article}

\usepackage[final]{acl}

\usepackage{times}
\usepackage{latexsym}

\usepackage[T1]{fontenc}
\usepackage[utf8]{inputenc}

\usepackage{microtype}

\usepackage[switch]{lineno}
\usepackage{amssymb}
\usepackage{latexsym}
\usepackage{mathtools}
\usepackage{algorithm}
\usepackage[noend]{algpseudocode}
\usepackage{setspace}

\usepackage{multirow}
\usepackage{tabularx}
\usepackage{threeparttable}
\usepackage{booktabs}
\usepackage{stfloats}
\usepackage{subfigure}
\usepackage{colortbl}
\usepackage{xcolor}
\usepackage{CJKutf8}

\title{Is MultiWOZ a Solved Task? \\
An Interactive TOD Evaluation Framework with User Simulator}

\author{Qinyuan Cheng\textsuperscript{1}\footnotemark[1], Linyang Li\textsuperscript{1}\footnotemark[1], Guofeng Quan\textsuperscript{1}, Feng Gao\textsuperscript{2}, Xiaofeng Mou\textsuperscript{2} and Xipeng Qiu\textsuperscript{1}\footnotemark[2] \\
\textsuperscript{1}School of Computer Science, Fudan University\\
\textsuperscript{2}AI Innovation Center, Midea Group Co Ltd \\
\{chengqy21, gfquan21\}@m.fudan.edu.cn \{linyangli19, xpqiu\}@fudan.edu.cn \\ \{gaofeng14, mouxf\}@midea.com
}

\begin{document}
\maketitle
\begin{CJK*}{UTF8}{gbsn}

\begin{abstract}

Task-Oriented Dialogue (TOD) systems are drawing more and more attention in recent studies.
Current methods focus on constructing pre-trained models or fine-tuning strategies while the evaluation of TOD is limited by a policy mismatch problem.
That is, during evaluation, the user utterances are from the annotated dataset while these utterances should interact with previous responses which can have many alternatives besides annotated texts.
Therefore, in this work, we propose an interactive evaluation framework for TOD. 
We first build a goal-oriented user simulator based on pre-trained models and then use the user simulator to interact with the dialogue system to generate dialogues.
Besides, we introduce a sentence-level and a session-level score to measure the sentence fluency and session coherence in the interactive evaluation. 
Experimental results show that RL-based TOD systems trained by our proposed user simulator can achieve nearly 98\% inform and success rates in the interactive evaluation of MultiWOZ dataset and the proposed scores measure the response quality besides the inform and success rates.
We are hoping that our work will encourage simulator-based interactive evaluations in the TOD task \footnote{\url{https://github.com/xiami2019/User-Simulator}}.

\renewcommand{\thefootnote}{\fnsymbol{footnote}} %将脚注符号设置为fnsymbol类型，即特殊符号表示
\footnotetext[1]{These authors contributed equally to this work.}
\footnotetext[2]{Corresponding author.}

\end{abstract}

 \vspace{0.001cm}

\section{Introduction}

Building intelligent dialogue systems has become a trend in natural language process applications especially with the help of powerful pre-trained models.
Specifically, task-oriented dialogue (TOD) systems \cite{DBLP:journals/corr/abs-2003-07490} are to help users with scenarios such as booking hotels or flights.
These TOD systems \cite{DBLP:conf/eacl/Rojas-BarahonaG17,DBLP:conf/acl/SocherZX18,DBLP:conf/acl/ChenCQYW19} usually first recognize user's intents and then generate corresponding responses based on an external database containing booking information.
Therefore, the key factor in TOD is the interaction between users and dialogue systems.

\begin{figure}[]
\centering
\includegraphics[width=1.0\linewidth]{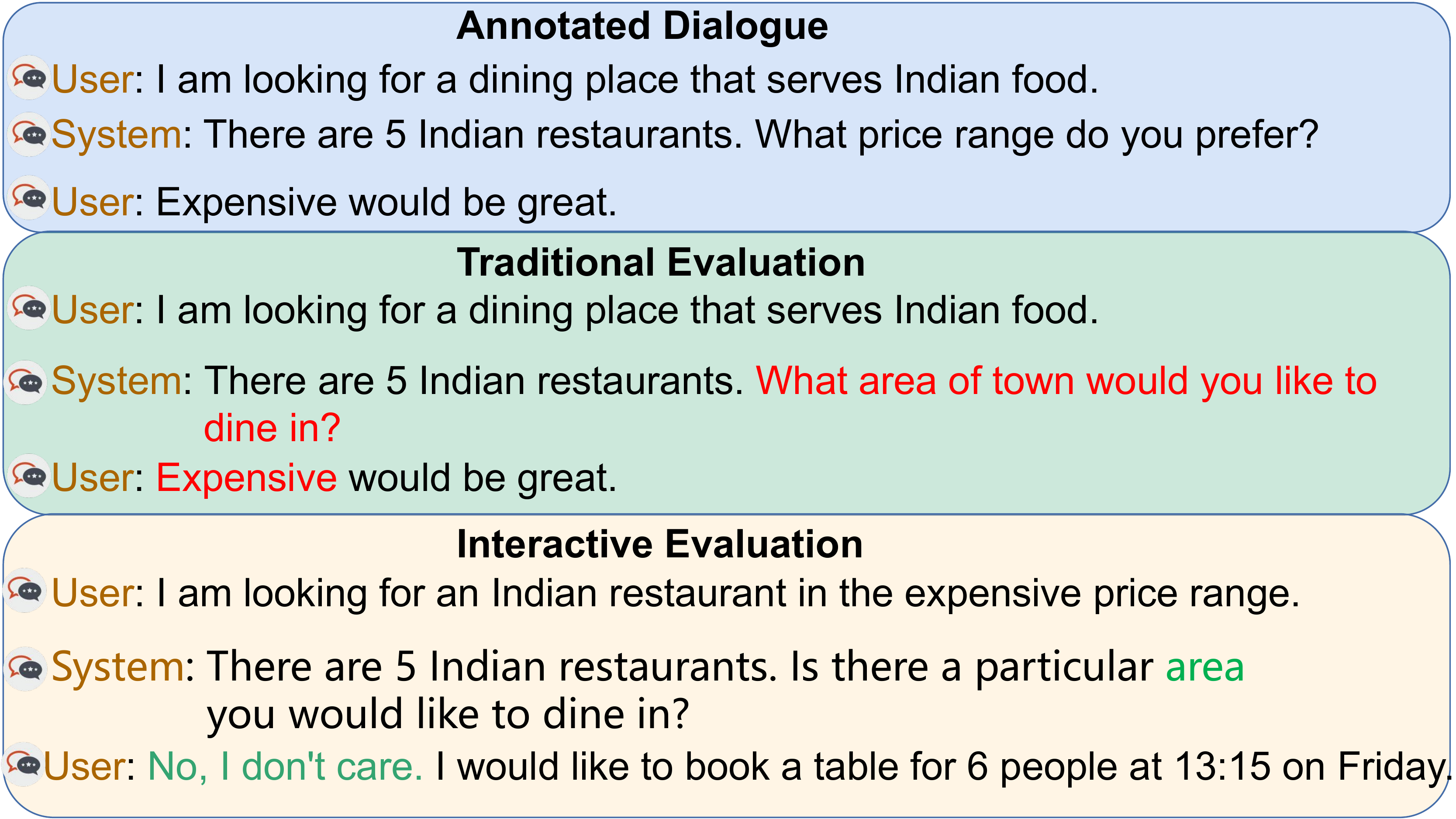}
\centering
\caption{Illustration of interactions between users and systems. Traditional evaluation might face a policy mismatch between utterances annotated with red color.}
\label{fig:illustration}
\end{figure}

However, in current TOD system evaluation process, traditional evaluation process uses annotated user utterances in multi-turn dialogue sessions no matter what responses the dialogue system generated, as illustrated in Figure \ref{fig:illustration}.
While in real-world dialogues, the user utterances are coherent with responses from the other speaker (which is the service provider).
Therefore, in TOD evaluation, using annotated utterances without interaction with the dialogue system will cause a policy mismatch, which would weaken the soundness of the evaluation results.
The mismatch might hurt the evaluation process since some responses may be correct and coherent but use a different policy with the annotated responses. 
Also, incoherent dialogue histories will affect the response generation.
With current state-of-the-art models achieving similar performance, 
% it is natural to consider that what current TOD needs is not better models but better evaluation strategies.
it is natural to consider that the bottleneck in the performance of current TOD systems is not the model capability but the evaluation strategy.
Since incorporating human interactions during evaluation is costly, a feasible method is to build an automatic interactive evaluation framework that can solve the policy mismatch problem.

In this paper, we propose a complete interactive evaluation framework to evaluate the TOD system.
We first build a strong dialogue user simulator based on pre-trained models, 
and we use the proposed simulator to deploy interactive evaluations.

In simulator learning, 
we introduce a goal-guided user utterance generation model based on sequence-to-sequence pre-trained models.
Then we use reinforcement learning to train both the user simulator and the dialogue system to boost interaction performance.

In interactive evaluations, we use the simulator to generate user utterances based on interactions with the generated responses instead of using static user utterances that have been annotated in advance.
Therefore, during evaluation, user utterances will respond to the generated responses, which can avoid the mismatch between stale user utterances and generated responses.
Further, in interactive evaluations, the quality of the generated texts cannot be measured since traditional BLEU cannot be calculated without oracle texts. 
To better evaluate the performance of dialogue systems,  
we introduce two automatic scores to evaluate the response quality at both sentence-level and session-level.
The sentence-level score is to evaluate the sentence fluency and the session-level score is to evaluate the coherence between turns in a dialogue session.
Also, these proposed scores can be used in traditional evaluation methods as well as the annotated dataset as a meta-evaluation to explore the importance of using user simulators to construct interactive evaluations.

We construct experiments on MultiWOZ dataset \cite{DBLP:conf/emnlp/BudzianowskiWTC18} based on pre-trained models and use our proposed simulator and scores to run interactive evaluations.
Experimental results show that interactive evaluations can achieve over 98\% inform and success rates, indicating that the bottleneck of TOD performance is the lack of proper evaluation methods. 
The proposed scores show that our proposed simulator can help achieve promising evaluation results in the interactive evaluation framework. 
Also, we explore the performance of RL-based models and we also use proposed scores to find that RL methods might hurt response quality to achieve high success rates.

Therefore, we can summarize our contributions:

% \begin{itemize}
    % \item 
    (A) We construct an evaluation framework that avoids policy mismatch problems in TOD.
    % \item
    
    (B) We build a strong user simulator for TOD systems that can be used in TOD training and evaluation.
    % \item 
    
    (C) Experimental results show the importance of using our proposed simulator and evaluation framework and provide hints for future TOD system developments with public available codes.
% \end{itemize}

\section{Related Work}

\subsection{Task-Oriented Dialogue Systems}

Task-oriented dialogue systems aim to achieve users' goals such as booking hotels or flights \cite{DBLP:conf/eacl/Rojas-BarahonaG17,DBLP:conf/sigdial/EricKCM17}.
% The general process of TOD can be divided into three subtasks: Dialogue State Tracking, Dialogue Policy Management and Response Generation. 
With the widespread use of pre-trained models \cite{DBLP:journals/corr/abs-2003-08271}, end-to-end TOD systems based on pre-trained models become more and more popular:
% \cite{DBLP:conf/nips/Hosseini-AslMWY20, DBLP:conf/aaai/YangLQ21, DBLP:journals/corr/abs-2109-14739,DBLP:conf/emnlp/Lee21,DBLP:journals/corr/abs-2111-14592}.
\citet{DBLP:conf/nips/Hosseini-AslMWY20} fine-tunes all subtasks of TOD using multi-task learning based on a single pre-trained model. \citet{DBLP:conf/aaai/YangLQ21} encodes results of intermediate subtasks, such as belief states and system actions, into dialogue history to boost responses generation. \citet{DBLP:journals/corr/abs-2109-14739} and \citet{DBLP:journals/corr/abs-2111-14592} use additional dialogue corpus to further pre-train the language model and then fine-tune the model on MultiWOZ dataset.  \citet{DBLP:conf/emnlp/Lee21} introduces an auxiliary task based on T5 models \cite{DBLP:journals/jmlr/RaffelSRLNMZLL20} and achieves state-of-the-art performance without using further pre-training methods.

\subsection{Automatic Evaluations}

Recent trends leverage neural models to automatically evaluate generated texts from different perspectives.
Automatic evaluation methods can help evaluate certain aspects in certain tasks such as factuality checking in text summarization \cite{DBLP:conf/emnlp/KryscinskiMXS20}, stronger BLEU score in machine translation \cite{DBLP:conf/acl/SellamDP20} and coherence in dialogue systems \cite{DBLP:conf/aaai/TaoMZY18,DBLP:conf/acl/PangNHZLT20}.
With pre-trained models, the quality of text generation can be measured by evaluation methods such as BERTScore \cite{DBLP:conf/iclr/ZhangKWWA20} and BARTScore \cite{DBLP:conf/nips/YuanNL21}.
With properly designed neural model scores, the performance of dialogue systems can be more accurately evaluated.

\subsection{User Simulators}

User simulators are designed to simulate users' behaviors in dialogue interactions, including rule-based simulators \cite{DBLP:conf/acl/LeeZTZZLLPLHG19} and model-based simulators \cite{DBLP:conf/acl/TakanobuLH20,DBLP:conf/acl/TsengDKB20}.
Usually, user simulators are introduced along with reinforcement learning strategies to enhance the dialogue policy modeling \cite{DBLP:journals/corr/LiLDLGC16,DBLP:conf/emnlp/ShiQWY19}, which can help the model learn better policies not included in the annotated data.
\citet{DBLP:conf/acl/TakanobuLH20} treats the model-based simulator as a dialogue agent like the dialogue system and formulate TOD as a multi-agent policy learning problem. \citet{DBLP:conf/acl/TsengDKB20} focuses on using reinforcement learning to jointly train the simulator and the dialogue system to boost the domain adaption capability of the model.

\section{Interactive Evaluation Framework}
% \section{Method}

In our proposed interactive evaluation framework, we first build a goal-state guided user simulator to model user policies and generate high-quality user utterances.
Then we construct the interactive evaluation framework and introduce two scores to evaluate the interactive inference results.

\subsection{User Simulator Construction}

A user simulator is to generate user utterances for interactions with the dialogue system.
Similar to the dialogue system construction, the user simulator also considers dialogue histories and generates utterances via a sequence-to-sequence text generation framework.
We propose a goal-state guided simulator that controls the user utterance generation based on the goal-state tracking.
Further, we adopt reinforcement learning methods to boost the interaction performance between our proposed goal-state guided simulator and dialogue systems.

\subsubsection{Goal-State Guided Simulator}

\begin{figure}[]
\centering
\includegraphics[width=1.0\linewidth]{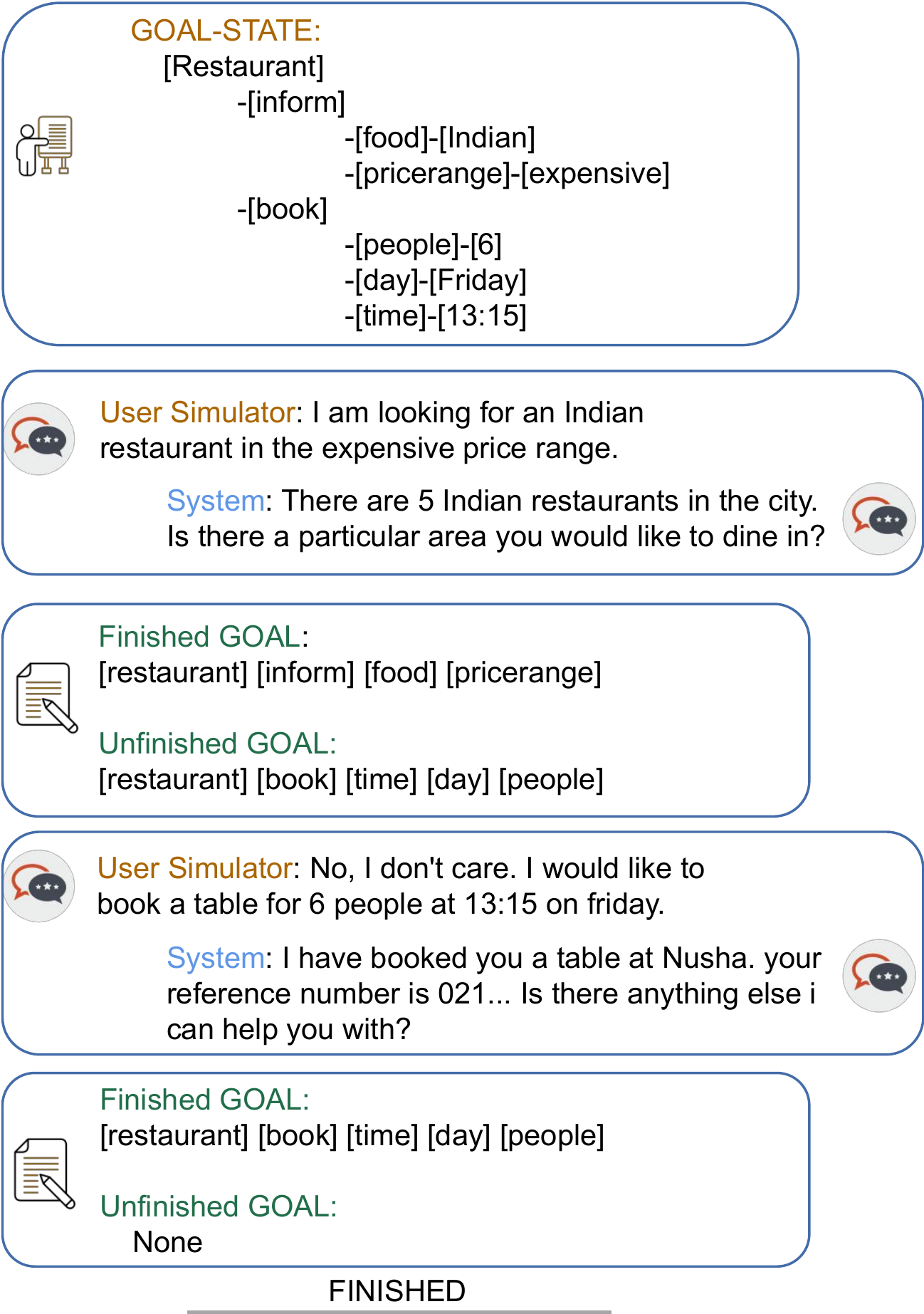}
\centering
\caption{Illustration of goal-state guided simulator interaction process, including goal states tracking and utterance generation.}
\label{fig:simulator}
\end{figure}

We introduce a goal-state guided simulator that generates user utterances based on sequence-to-sequence pre-trained models.
The basic idea is to use pre-defined user's goals as initial goal states and track goal states based on user and system actions, which is similar with belief state tracking.
As seen in Figure \ref{fig:simulator}, we illustrate the interaction process of the user simulator and the dialogue system.
We first add current goal states at the front of the user simulator inputs.
Plus, the user simulator will encode previous dialogue histories including user utterances and dialogue system responses.
The user simulator will predict the user actions and then obtain finished goals by combining both user actions and dialogue system actions.
By cutting off finished goals from the current goal states, we obtain unfinished goals and the user simulator will generate user utterances based on these unfinished goals at next turn.
When the user simulator has finished all the required goals, the unfinished goal slot is empty, the user simulator will cease to generate utterances.
Besides, we add two additional terminate signals for the user simulator.
When the dialogue session exceeds a certain number of turns and the goal states still cannot be fully finished or when the user simulator or the dialogue system generates definite actions to stop the session like action 'bye' or action 'thank', the user simulator will terminate the dialogue session.

\subsubsection{Simulator Training}

The training process of the user simulator includes sequence-to-sequence supervised learning and reinforcement learning.

In supervised learning, the user simulator encodes the goal states at the front of the input texts and considers all dialogue histories including texts of both user utterances and system responses.
The generation texts include current user actions and user utterances.
Therefore, the entire training process is a standard sequence-to-sequence generation task training process optimized by cross entropy loss.

In addition, we incorporate a reinforcement learning strategy to further boost the interaction performance between user simulators and dialogue systems.
It is intuitive to use the inform and success rates as rewards to increase the interaction policies between simulators and dialogue systems.  
That is, supervised learning methods can only learn policies annotated in the training dataset, while reinforcement learning methods can bring new interaction policies for both simulators and dialogue systems.

Therefore, following policy gradient algorithm \cite{DBLP:conf/nips/SuttonMSM99}, we use the success rate to construct rewards to optimize both user simulator $\theta$ and dialogue system $\phi$.
For each turn $t$, we consider the process of generating each token as an action.
Therefore, we calculate turn-level RL gradients $\mathcal{J}$ based on each generated token $i$ in user simulator and $j$ in dialogue system:

\begin{align}
   \nabla_{\theta} \mathcal{J}(\theta) = \sum_{i=1} \gamma^{|A_t|-i} R_t \nabla_{\theta} \mathcal{\pi}_{}(\theta,i) \\
   \nabla_{\phi} \mathcal{J}(\phi) = \sum_{j=1} \gamma^{|A_t|-j} R_t  \nabla_{\phi} \mathcal{\pi}_{}(\phi,j)
\end{align}

Here, $\gamma$ is the discounting factor and $|A_t|$ is the policy sequence length of turn t, $\mathcal{\pi(\cdot)}$ is the forward strategy of corresponding simulator $\theta$ and dialogue system $\phi$.
The reward $R_t$ is 1 when the generated session from simulator and dialogue system interaction successfully achieves user's goals and 0 otherwise.

\subsection{Evaluation Framework Construction}

Current TOD evaluation strategies will face a policy mismatch between users and dialogue systems.
Therefore it is necessary to construct a complete evaluation framework based on simulator interactions to avoid the policy mismatch when evaluating TOD systems.
Based on the proposed simulator, we evaluate TOD systems interactively.
Further, we propose sentence-level and session-level scores in the interactive evaluation process to measure the quality of generated responses.
Plus, we apply the introduced scores for RL training to further explore the dialogue system performance.

\subsubsection{Traditional Evaluation}

Traditional evaluation process to test TOD systems is to use inform and success rates to measure whether the responses achieve user's goals in the given session.
During the evaluation process, the responses are generated from the dialogue systems but the user utterances are from the annotated dataset regardless of the interactions between responses and the following user utterances.
Therefore, the whole session can be unnatural due to the policy mismatch problem.
Further, BLEU score is used to measure the quality of the generated responses compared with the annotated oracle responses.
The annotated responses may be entirely different from the generated responses when there is a policy mismatch.
For instance, in the MultiWOZ dataset, responses generated by MTTOD \cite{DBLP:conf/emnlp/Lee21} model achieve an average BLEU score of 19.47, yet 4686 of 7372 responses obtain BLEU scores below 1.  
% It is intuitive that the cause of the low BLEU scores might not be the problem of the dialogue system since the average BLEU is quite high.
Such a phenomenon that the responses can achieve either high or nearly zero BLEU scores indicates that the evaluation process might suffer from policy mismatch problems.
Therefore, we believe that the bottleneck of current TOD systems is the stale evaluation process with policy mismatches.

\subsubsection{Interactive Evaluation}

Interactive evaluation is used to evaluate the dialogue responses based on interactions between users and systems.
Therefore, the user utterances can interact with the dialogue responses, which can avoid evaluation errors when the responses are reasonable but do not match the following utterances.
During interactive evaluation, we suppose that the user simulator can produce high-quality user utterances based on the pre-defined goals the user aimed to achieve.
The evaluation metrics inform and success rates are the same with traditional evaluation.

\subsubsection{Sentence-Level Score}

During the interactive evaluation, the quality of the generated response cannot be measured by BLEU score.
To properly evaluate the quality of the responses generated during interactions with the user simulator, we use an automatic score to evaluate the quality of the generated responses.

Specifically, similar to text perplexity and BARTScore evaluation \cite{DBLP:conf/nips/YuanNL21}, we measure the sentence fluency by using the score calculated as:

\begin{align}
    S = -\sum_{i=1}^{L} \frac{1}{L}log p (\boldsymbol{y}_i|\boldsymbol{y}_{<i}, \theta) 
\end{align}
Here, $\boldsymbol{y}_i$ denotes the $i^{th}$ token in the generated response; $\theta$ represents the language model, which is a fine-tuned GPT-2\cite{Radford2019LanguageMA} in our experiments; $\boldsymbol{L}$ is the sequence length.

With such a score, we can properly measure the response quality when the BLEU score is no longer applicable as the metric for dialogue response quality evaluation.

\subsubsection{Session-Level Score}

Besides sentence-level response quality, it is also important to evaluate the coherence between the dialogue system and the user simulator.
Therefore, we propose a session-level score to explore the coherence between user utterances and dialogue responses.

Inspired by the sentence-relation tasks such as natural language inference task \cite{maccartney-manning-2008-modeling} and next sentence prediction task \cite{Devlin2019BERTPO}, we construct a binary classification model to predict whether the interactions between utterances and responses are coherent and fluent.
Naturally, the user utterance and the response pair are usually coherent since the dialogue system is trained to answer the user's queries.
The response might not be coherent with following-up user utterances in the traditional evaluation process. 
Therefore, we consider a continuous session including one user utterance denoted as $\boldsymbol{u}_{t}$ and its corresponding response $\boldsymbol{r}_{t}$, plus the next user utterance $\boldsymbol{u}_{t+1}$.
We suppose that dialogue turns such as $[\boldsymbol{u}_{t}, \boldsymbol{r}_{t} ]$ and $[\boldsymbol{r}_{t}, \boldsymbol{u}_{t+1} ]$ are coherent in nature.
We select a random response $\boldsymbol{r}_{}^*$ to construct negative samples to train the binary classifier to score the session-level coherence to evaluate the interaction between user utterances and system responses.
After building the binary classifier, we use the average score (softmaxed confidence) of all $[\boldsymbol{u}_{t}, \boldsymbol{r}_{t} ]$ and $[\boldsymbol{r}_{t}, \boldsymbol{u}_{t+1} ]$ pairs in a session as the session-score.

The session-level score can be used as an evaluator that measures the entire session fluency, that is, during the interaction process, the session-level scorer evaluates the overall quality of both user utterances and dialogue responses.
Therefore, such a score can be used in evaluating both user simulator performance and dialogue system performance.

\section{Experiments}

\subsection{Datasets and Implementations}

\subsubsection{Datasets}

We use MultiWOZ 2.0 dataset \cite{DBLP:conf/emnlp/BudzianowskiWTC18} to construct all our experiments.
MultiWOZ 2.0 contains 10,438 sessions with 115,434 turns in 7 different domains.
In MultiWOZ dataset, the initial goal state information is given, which is the information originally provided for annotators to craft the MultiWOZ dataset.
Therefore, these goal states are suitable for the simulator training.

\subsubsection{Dialogue System Implementations}

In the interactive evaluation process, we use several state-of-the-art TOD systems including UBAR \cite{DBLP:conf/aaai/YangLQ21}, PPTOD \cite{DBLP:journals/corr/abs-2109-14739} and MTTOD \cite{DBLP:conf/emnlp/Lee21}.
For a fair comparison, we re-implement all these models and adapt the same data processing and evaluation scripts.
For the MTTOD model, we use a T5-small as the backbone and remove its auxiliary task and additional decoder for the simplicity of the experiment, which causes no performance degeneration.
% The hyper-parameters of these models follow their original paper. 
For the PPTOD model, we also use its small version. 
Our code is based on huggingface transformers and the MTTOD implementations.

For the RL-based model, we initialize model parameters from a supervised learning based model and use a Monte Carlo based policy gradient method which is time consuming.
Therefore, in each RL epoch, we random pick 200 user's goals from training set (corresponding to 200 sessions from training set) to do 400 episodes for reinforcement learning. 
And during training, we first fix one agent and update the other one for 200 episodes and then vice versa.

\subsubsection{Simulator Implementations}

We implement the simulator based on T5 \cite{DBLP:journals/jmlr/RaffelSRLNMZLL20}, specifically the small and base version.
The supervised training process of the simulator is similar to dialogue system training using similar hyper-parameters.
For the reinforcement learning process, we use different random seeds to construct multiple runs.

\subsubsection{Score Implementations}

For the sentence-level score, we use a GPT-2 model and fine-tune the model using the MultiWOZ dataset for the special token learning.
For the session-level score, we use a BERT-base model as the binary classification model and use the average softmax logits.

\begin{table}[]
    % \scriptsize
    \small
    \centering
    \begin{tabular}{ll}
    \toprule
        \multicolumn{1}{c}{\bfseries Simulator Backbone}  & \multicolumn{1}{c}{\bfseries BLEU} \\
        \cline{0-1}
        T5-Base & 17.72 \\
        % \cline{0-1}
        T5-small & \textbf{17.92} \\
        \cline{0-1}
    \end{tabular}
    \caption{User simulator performance}
    \label{tab:user-simulator}
\end{table}

\subsection{User Simulator Performances}

As seen in Table \ref{tab:user-simulator}, we first test the simulator quality based on the traditional evaluation, specifically BLEU score using oracle user utterances as references.
For simplicity, we use oracle dialogue histories as the inputs of the simulator. And the BLEU score reflects that our simulator could generate natural utterances.
Since the T5-small model achieves a higher BLEU score, we use it for all following experiments.

\begin{table*}[]\setlength{\tabcolsep}{8pt}
    \scriptsize
    % \small
    \centering   
    % \resizebox{\textwidth}{!}{
    \begin{tabular}{lccccccccccccccccccc}
        \toprule

      \multicolumn{2}{c}{\bfseries Models} & \multicolumn{6}{c}{\bfseries Evaluation Metric}  \\ 
      Dialogue System & Simulator  &  Inform & Success & BLEU & Comb. Score & Sent-Score $\downarrow$ & Sess-Score $\uparrow$ \\
        \midrule
        \midrule
       \multicolumn{1}{c}{\bfseries Traditional Evaluation} \\
        \midrule
        UBAR \cite{DBLP:conf/aaai/YangLQ21} & N/A & 91.90 & 82.80 & 19.10 & 106.45 & 1.11 & 82.7 \\
        PPTOD \cite{DBLP:journals/corr/abs-2109-14739} & N/A & 92.00 & 79.90 & 18.42 & 104.37 & 1.14 & 82.8 \\
        MTTOD \cite{DBLP:conf/emnlp/Lee21} & N/A & 94.30 & 85.40 & 19.47 & 109.32 & 1.09 & 83.7 \\

        \midrule 
        \multicolumn{1}{c}{\bfseries Interactive Evaluation} \\
        \midrule
        UBAR  & T5 & 95.10 & 92.00 & N/A & N/A & \text{0.80} & \text{97.0}  \\
        PPTOD & T5 & 91.50 & 86.20  & N/A & N/A &  0.83 & 95.9 \\
        MTTOD & T5 & 94.00 & 91.00 & N/A & N/A & \textbf{0.79} & \textbf{97.1}  \\
        MTTOD (RL) & T5 (RL) & \textbf{98.04} & \textbf{97.10} & N/A & N/A & 0.87 & 93.8 \\

        \midrule 
        \multicolumn{1}{c}{\bfseries Meta Evaluation} \\
        \midrule        
        
        Testset & N/A & N/A & N/A &N/A  & N/A & 1.44 & 93.1 \\

        \bottomrule

    \end{tabular}
    \caption{Results of different models under different evaluation framework}
    \label{tab:main-combine-task}
\end{table*}

\subsection{Interactive Evaluation Experiment}

\subsubsection{Inform and Success Rates}

As seen in Table \ref{tab:main-combine-task}, on the same model (such as MTTOD), the success rate of using traditional evaluation is much lower compared with using interactive evaluation (85.40\% compared to 91.00\%). 
We can observe that when the dialogue system and the user simulator are both optimized by RL algorithm, the inform and success rates can reach nearly 98\%, indicating that with our proposed user simulator, the user's goals can be well accomplished by the dialogue system.
Therefore, with pre-trained models, multi-task learning and reinforcement learning using a strong user simulator, 
\textbf{is MultiWOZ a solved dataset already?} 
We might begin to consider such a possibility.
The incredibly high inform and success rates using interactive evaluation compared with traditional evaluation results indicate that the bottleneck that constrains previous evaluation performance might not be the deficiency of the dialogue system but the policy mismatch problem in traditional evaluation methods. 

\subsubsection{Different Models}

We use both traditional and interactive evaluation methods to compare the performance of several state-of-the-art models.
As seen in Table \ref{tab:main-combine-task}, the MTTOD model achieves the highest inform and success rates in the traditional evaluation while the UBAR model achieves the highest in the interactive evaluation, indicating that the improvements between these state-of-the-art models are not large enough to create significant difference.
Meanwhile, results of the session-score show that the generated dialogues in traditional evaluations are incoherent, which is caused by the policy mismatch problem. 
The generated dialogues in interactive evaluations are more coherent.
We can observe that the results of interactive evaluations are better than the results of traditional evaluations on all models. 
This result shows that policy mismatch problems in traditional evaluations do limit model's performance.

\subsection{Sentence-Score Evaluation}

In the interactive evaluation framework, the fluency of the generated responses cannot be properly measured since traditional BLEU score is not available.
With the sentence-score, we can measure the sentence fluency without reference texts.
As seen in Table \ref{tab:main-combine-task}, responses generated in the interactive evaluation obtain the best performance.
Compared with the oracle test set, sentence-scores of the interactive evaluation are even better. 
This is because the generated utterances from models trained by maximum likelihood estimation prefer to use the patterns that appear more frequently in the training set, therefore the cross entropy loss of these utterances will be lower. 
However, in the oracle test set, utterances are more complicated and diversity caused by different annotators, which may lead to higher cross entropy loss then generated utterances.
Therefore, sentence-scores of generated utterances are better than oracle utterances from test set.
Besides, we can observe that RL-based models obtain worse sentence-scores compared with supervised learning based models (0.87 compared to 0.79).
RL-based models achieving promising inform and success rates yet worse sentence-scores indicates that using reinforcement learning methods to optimize inform and success rates will cause the degeneration of sentence quality.
Through such an observation of the sentence-score in our interactive evaluation framework, we can conclude that RL-based models might hurt the quality of the generated utterances.

\begin{table*}[]\setlength{\tabcolsep}{8pt}
    \scriptsize
    \centering  
    % \resizebox{\textwidth}{!}{
    \begin{tabular}{llcccccccccccccccccc}
        \toprule

      \multicolumn{2}{c}{\bfseries Models} & & \multicolumn{5}{c}{\bfseries Evaluation Metric}  \\ 
      Dialogue System & Simulator & Rand. Seed &  Inform & Success  & Sent-Score ($\downarrow$) & Sess-Score ($\uparrow$) & \\
        \midrule

        \multicolumn{1}{c}{\bfseries Interactive Evaluation} \\
        \midrule

        MTTOD & T5 & - &  94.00 & 91.00  & \textbf{0.79} & \text{97.1} \\
        MTTOD (RL) & T5 (RL) & avg. & \textbf{98.04} & \textbf{97.10}  & 0.87 & 93.8 \\
        MTTOD (RL-Sess) & T5 (RL-Sess) & 1997 & 93.70 & 92.40 & 0.81 ($\downarrow$ 0.06) & \text{96.8} ($\uparrow$ 3.0) \\
        MTTOD (RL-Sess) & T5 (RL-Sess) & 1998 & 93.40 & 89.20 & 0.83 ($\downarrow$ 0.04) & \text{95.0} ($\uparrow$ 1.2) \\
        
        MTTOD (RL-Sent) & T5 (RL-Sent) & 1997 & 96.80 & 96.40 & \text{0.81} ($\downarrow$ 0.06) & 97.3 ($\uparrow$ 3.5) \\
        MTTOD (RL-Sent) & T5 (RL-Sent) & 1998 & 92.80 & 88.40 & 0.81 ($\downarrow$ 0.06) & \textbf{97.4} ($\uparrow$ 3.6)  \\

        \bottomrule

    \end{tabular}
    \caption{Results of different RL settings}
    \label{tab:rl-results}
\end{table*}

\subsection{Session-Score Evaluation}

It is also important to consider session-level interaction quality in the interactive evaluation framework.
Therefore, we construct the session-score to measure the coherence between user utterances and system responses.
As seen, traditional evaluations achieve relatively low session-scores (about 83\%), indicating that the user utterances and the system responses in a single session are incoherent.
As for the interactive evaluation process, session-scores are considerable high, indicating that sessions generated by interactions are coherent.
Also, we can use the session-score as a meta evaluation to measure the test set sessions.
As seen, test set session-score achieves a significantly higher score compared with the sessions generated in the traditional evaluation process.
As for the comparison between sessions in the interactive evaluation and the test set, we can observe that the sessions from interactions achieve higher session-scores (97.1\% compared to 93.1\%).
We assume that this is because the annotated test set contains more variance in session construction, which cannot be completely understood by neural-based scores.
Through session-score evaluations, we can conclude that when using the interactive evaluation to test the TOD system, the dialogue session is more natural compared with using annotated user utterances and generated responses in the traditional evaluation process. Besides, we can observe that using reinforcement learning will slightly hurt the session coherence(93.8\% compared to 97.1\%).

\subsection{RL Based Model Performance}

We run 5 times for each RL-based model with different random seeds and show the variance in Appendix, since RL methods are unstable. 

\textbf{RL-Method Results}

As seen in Table \ref{tab:main-combine-task}, with RL training, the average inform and success rates achieve a significant improvement while the sentence-score and the session-score grow lower.
These results indicate that though the inform and success rates get better, the text quality might be worse.
While RL-based methods are widely used in joint training of dialogue systems and user simulators at present, we raise concerns about using the success rate as the reward function will hurt the quality of generated utterances considering our experiment results.

\textbf{Score Results as Rewards}

Further, since we introduce new scores to evaluate the dialogue system, it is intuitive to utilize these scores as rewards to improve the RL training of the dialogue system.
Therefore, in Table \ref{tab:rl-results}, we conduct experiments to explore whether using these scores as rewards is helpful.
We design two new reward functions: RL-Sent and RL-Sess.
\begin{align}
    R = \text{Success} + \alpha * \frac{1}{\text{Sent}} + \beta * \text{Sess}
\end{align}
When $\alpha$ is 0.1 and $\beta$ is 0,  the reward setting is RL-Sent;when $\alpha$ is 0 and $\beta$ is 0.1, the reward setting is RL-Sess.
\footnote{$\alpha$ and $\beta$ are modifiable hyperparameters.}

As seen, adding sentence-score or session-score to the rewards to train the dialogue system can obtain better results on the corresponding score.
Still, using these rewards in RL causes more instability that some seeds may achieve promising results given these scores as rewards but some seeds may reach frustrating performance.
We conclude that using these scores as rewards to improve the quality of the generated responses is a huge challenge that requires more careful design of RL-algorithm.

\begin{table*}[]
\centering
\fontsize{8}{9}\selectfont
\begin{tabular}{lllll}
% \cline{1-4}
\toprule
Model & Text & Sent.$\downarrow$ & Sess.$\uparrow$ &  \\
\cline{0-4}
\multirow{1}{*}{\begin{tabular}[c]{@{}l@{}} T5 Simulator \\ MTTOD \\ \end{tabular}} &
% \begin{tabular}[c]{@{}l@{}}User: Can I get the postcode, phone number, and if they have internet?  \\  \color{teal}{System: Peace Hotel has free internet and free parking. Their phone number is 021...,} \\\color{teal}{ and their postcode is 9....}\\
% User: I also need a train to Ely on Friday.
% \end{tabular} 
\begin{tabular}[c]{@{}l@{}}User: I am also looking for a restaurant that serves french food and is expensive. \\  \color{teal}{System: There are 3 restaurants that meet your criteria. Would you like me to book you a table?} \\
User: No, I just need the postcode and phone number. \\
\color{teal}{System: The phone number is ... and the postcode is ... Is there anything I can help you with?}
\end{tabular} 
& \begin{tabular}[c]{@{}l@{}} \\ 0.81 \\ \\ 0.34 \end{tabular} & 98.7 & \\ 
\cline{0-4}
\multirow{1}{*}{\begin{tabular}[c]{@{}l@{}} T5 Simulator \\ MTTOD(RL) \end{tabular}} &
% \begin{tabular}[c]{@{}l@{}}User: Can you give me the postcode and phone number?\\ \color{teal}{System: Peace Hotel is} \color{red}{located at postcode 9....}\color{teal}{The phone number is 021-...}
% \\User: Do they have internet?\\ \color{teal}{System: Yes, Peace Hotel has Internet.}\\
% User: I am also looking for a train to Ely on Friday.\end{tabular} 
\begin{tabular}[c]{@{}l@{}}User: I am also looking for an expensive restaurant.\\ \color{teal}{System: There are 3 options. Is there a particular area you would like to dine in?}
\\User: No , I don't care. Can you give me the postcode and phone number?\\ \color{teal}{System: } \color{red}{Cote is cb21uf.} \color{teal}{Their phone number is 213-...}\end{tabular} 
& \begin{tabular}[c]{@{}l@{}} \\ 0.72 \\ \\ 1.49 \end{tabular} & 98.6 &  \\ 
\cline{0-4}
\multirow{1}{*}{\begin{tabular}[c]{@{}l@{}} Traditional Eval \\ MTTOD \\ \end{tabular}} & \begin{tabular}[c]{@{}l@{}}User: I need to book a taxi please.\\  \color{teal}{System: I would be happy to help! Where will you be departing from?}\\
\color{red}{User: I want to arrive by 12:45.}\\
\color{teal}{System: Where will you be going?}
\end{tabular} & \begin{tabular}[c]{@{}l@{}} \\ 0.86 \\ \\ 0.65 \end{tabular} & 57.2 &  \\  
\cline{0-4}
\multirow{1}{*}{\begin{tabular}[c]{@{}l@{}} TestSet \\  \\ \end{tabular}} &
% \begin{tabular}[c]{@{}l@{}}User: How about the moderate one? May I have their address, please? \\   \color{teal}{System: Yes Royal Standard 's address is Royal Street... and their postcode is 9...} \\ \color{teal}{Is there anything else I can help you with today?}\\
% \color{black}{User: No, that is all, thank you. Have a nice day .}\\
% \color{teal}{System: So glad we could help you out. Thanks for using the Cambridge Towninfo centre, }\\
% \color{red}{and have a glorious day!}
% \end{tabular} 
\begin{tabular}[c]{@{}l@{}}User: I need a taxi to pick me up at Regency Gallery and take me to Don Pasquale Pizzeria. \\   \color{teal}{System: What time would you like to arrive there by?} \\
\color{black}{User: I would like to arrive by 20:00.}\\
\color{teal}{System: } \color{red}{Okay , you're all set. Be on the look out for a blue Tesla.} \color{teal}{You can reach the taxi at ...} \\
\color{teal}{Can I do anything more for you?}\\
\end{tabular} 
& \begin{tabular}[c]{@{}l@{}} 1.16 \\ \\ 1.38 \\ \end{tabular} & 98.0 &  \\

\bottomrule
%  &  &  &  &  \\
%  &  &  &  & 
\end{tabular}
\caption{Case studies in the MultiWOZ dataset with sentence and session scores.}
\label{tab:casestudy}
\end{table*}

\subsection{Case Studies}

\subsubsection{RL-Enhanced Responses}

The responses generated by the RL-based models can be sometimes stale in format and focus on key values that might help improve inform and success rates.
As seen in the second group of Table \ref{tab:casestudy}, the system responses include inarticulate phrases such as \textit{cote is cb21uf} (cote is the name of the restaurant and cb21uf is its postcode) since the key values postcode and name are important for calculating inform and success metrics.
The sentence-score also gives a worse score for this utterance. 
Such a phenomenon indicates that incorporating RL algorithm in training dialogue systems and simulators requires further attention to maintaining high response quality besides inform and success rates.

\subsubsection{Session-Level Coherence}

As seen in Table \ref{tab:casestudy}, the session-score predicts a lower score when evaluating the session from the traditional evaluation.
The utterance \textit{I want to arrive by 12:45.} cannot match the dialogue response querying the departure place.
Such a result indicates that sessions in the traditional evaluation process can be unnatural which may constrain further improvements of TOD system developments.
Also, high session-scores of sessions generated in the interactive evaluation process indicate that such an evaluation process is more natural therefore can be a more appropriate evaluation standard for TOD systems.

\subsubsection{Diversity in the Testset}

Further, since we observe that the testset sessions achieve rather poor performance in the sentence-score results, we assume that the diversity in the testset brings difficulties for the sentence-score and session-score.
Therefore, we explore cases in the testset and we find that as seen in Table \ref{tab:casestudy}, human-wrote system responses contain a large proportion of diversified responses such as \textit{be on the look out for}.
These patterns are common in the testset sessions and we do \textbf{not} cherry pick a bad sentence-score system response case.
Therefore, we can summarize that our proposed sentence-score can make fair comparisons between model generated responses while neural model based scores can still be improved. 

\section{Conclusion}

In this paper, we focus on the evaluation of current end-to-end TOD systems.
% We introduce a strong user simulator and use it to run interactions with dialogue systems to evaluate current state-of-the-art TOD systems.
We construct an automatic interactive evaluation framework with a strong user simulator.
Besides, we obtain extremely high interactive evaluation performance on MultiWOZ dataset by jointly training our user simulator and dialogue system.
Through the interactive evaluation framework, we can conclude several hints for future studies on TOD systems:
(1) current TOD needs more challenging and complicated dataset and scenarios;
(2) interactive evaluation process should be considered in proper evaluation of TOD systems;
(3) reinforcement learning used in training user simulators and dialogue systems requires more careful design to consider both task success rate and quality of generated texts.

\section{Limitations}
Based on our experiments and conclusions, we conclude some limitations of our work as follows: 
\begin{itemize}
    \item One limitation of our work is that, given the current datasets used in TOD, we can only conclude that the datasets are well-solved under the interactive evaluation setting without exploring more challenging dataset settings that have more challenging scenarios and evaluation perspectives.
    \item In training our proposed sentence-score and session-score, these neural model-based scores are trained by maximum likelihood estimation(MLE), which intend to give better scores to patterns that appear more frequently in training data. 
    Such a limitation is a challenge in current neural network based automatic scores which can be further explored in not only the dialogue area.
    \item For the design of the reward function, our method using a weighted sum of success rates, sentence-scores and session-scores is a straightforward method to maintain both high task success rate and generation quality.
    The design of RL-based methods can be further improved given multiple rewards from different perspectives in order to train a better dialogue system, which could be a potential future work.
\end{itemize}

\section*{Acknowledgement}
We would like to thank all anonymous reviewers for their valuable advice.
This work was supported by the National Key Research and Development Program of China (No.2020AAA0108702) and National Natural Science Foundation of China (No.62022027). 

\end{CJK*}

% \newpage

\bibliography{anthology}
\bibliographystyle{acl_natbib}

\appendix
\setcounter{table}{0}   %从0开始编号，显示出来表会A1开始编号
\renewcommand{\thetable}{A\arabic{table}}

\clearpage

\section*{Appendix}
\section{Different random seeds for RL training}
The results of different random seeds and different RL settings are shown at Table \ref{tab:rl-results-all}.
As seen, RL-based models can get good performance when only using success rates as rewards. However, when using RL-Sess or RL-Sent, some runs may encounter unsatisfied results, indicating that models are unstable during training.

\begin{table*}[]\setlength{\tabcolsep}{5pt}
    \centering   
    \begin{tabular}{llcccccccccccccccccc}
        \toprule

      \multicolumn{2}{c}{\bfseries Models} & & \multicolumn{5}{c}{\bfseries Evaluation Metric}  \\ 
      Dialogue System & Simulator & Rand. Seed &  Inform & Success  & Sent-Score ($\downarrow$) & Sess-Score ($\uparrow$) & \\
        \midrule

        \multicolumn{1}{c}{\bfseries Interactive Evaluation} \\
        \midrule

        MTTOD & T5 & - &  94.00 & 91.00  & \textbf{0.79} & \text{97.1} \\
        MTTOD (RL) & T5 (RL) & 1996 & 98.20 & 96.40 & 0.94 & 92.0 \\
        MTTOD (RL) & T5 (RL) & 1997 & 98.40 & 98.00 & 0.88 & 90.3 \\
        MTTOD (RL) & T5 (RL) & 1998 & \textbf{98.90} & \textbf{98.50} & \textbf{0.79} & 95.7 \\
        MTTOD (RL) & T5 (RL) & 1999 & 97.40 & 96.90 & 0.85 & 95.5 \\
        MTTOD (RL) & T5 (RL) & 2000 & 97.30 & 95.70 & 0.88 & 95.6 \\
        \hline
        
        MTTOD (RL-Sess) & T5 (RL-Sess) & 1996 & 95.90 & 91.30 & 0.91 & 95.3 \\
        MTTOD (RL-Sess) & T5 (RL-Sess) & 1997 & 93.70 & 92.40 & 0.81 & 96.8 \\
        MTTOD (RL-Sess) & T5 (RL-Sess) & 1998 & 93.40 & 89.20 & 0.83 & 95.0 \\
        MTTOD (RL-Sess) & T5 (RL-Sess) & 1999 & 85.10 & 84.00 & 0.90 & 87.2\\
        MTTOD (RL-Sess) & T5 (RL-Sess) & 2000 & 96.70 & 86.90 & 0.80 & 93.4\\
        \hline
        
        MTTOD (RL-Sent) & T5 (RL-Sent) & 1996 & 97.00 & 96.90 & 0.88 & 94.8 \\
        MTTOD (RL-Sent) & T5 (RL-Sent) & 1997 & 96.80 & 96.40 & 0.81 & 97.3 \\
        MTTOD (RL-Sent) & T5 (RL-Sent) & 1998 & 92.80 & 88.40 & 0.81 & \textbf{97.4} \\
        MTTOD (RL-Sent) & T5 (RL-Sent) & 1999 & 83.80 & 79.60 & 0.84 & 86.2 \\
        MTTOD (RL-Sent) & T5 (RL-Sent) & 2000 & 96.90 & 92.70 & 0.81 & 93.8 \\

        \bottomrule

    \end{tabular}
    \caption{Results of different random seeds under different RL settings}
    \label{tab:rl-results-all}
\end{table*}

\section{Training Details}
\subsection{Devices}
We run all our experiments on a single RTX3090 with 24G gpu memory, including supervised training and reinforcement training.
\subsection{Hyperparameters}
\noindent\textbf{Supervised Learning}  

We train 20 epochs with batch size 8 for both our MTTOD model and user simulator. We train 20 epochs with batch size 32 for our PPTOD model. We train 40 epochs with batch size 4 for our UBAR model.  

\noindent\textbf{Reinforcement Learning}  

We only apply reinforcement learning for the MTTOD model and the user Simulator. We train 20 epochs for RL models. Each epoch contains 400 episodes. At first 200 episodes, we fix the dialogue model and update the user simulator. At last 200 episodes, we fix the user simulator and update the dialogue model. We use a half of the development set to select the best model due to the interaction is time-consuming.

\noindent\textbf{Score Training}  

We train 20 epochs with batch size 32 for both sentence-score and session score.

\end{document}